\documentclass{article}

\usepackage{microtype}
\usepackage{graphicx}
\usepackage{subfigure}
\usepackage{booktabs} % for professional tables
\usepackage{amsmath}
\usepackage{amssymb}
\usepackage{amsthm}
\usepackage{stackengine}
\usepackage{forest}
\tikzset{
  dot/.style={circle,draw,inner sep=1.2,fill=black},
}

\newtheorem*{observation}{Observation}

\newtheorem*{definition}{Definition}

\DeclareMathOperator*{\EE}{\mathbb{E}}

\newcommand{\KL}{\mathrm{KL}}

\newcommand{\objKL}{\mathcal{J}}
\newcommand{\pp}{\pi}
\newcommand{\qp}{\mu}
\newcommand{\ppi}{{\pi^{(i)}}}
\newcommand{\qpi}{{\mu^{(i)}}}
\newcommand{\ppnext}{{\pi^{(i+1)}}}

\newcommand{\ourmethod}{TreePI~}

\usepackage{hyperref}
\usepackage{nicefrac}

\usepackage[accepted]{icml2020}

\icmltitlerunning{Local Search for Policy Iteration in Continuous Control}

\begin{document}

\twocolumn[
\icmltitle{Local Search for Policy Iteration in Continuous Control}

\icmlsetsymbol{equal}{*}

\begin{icmlauthorlist}
\icmlauthor{Jost Tobias Springenberg}{equal,goo}
\icmlauthor{Nicolas Heess}{equal,goo}
\icmlauthor{Daniel Mankowitz}{goo}
\icmlauthor{Josh Merel}{goo}
\icmlauthor{Arunkumar Byravan}{goo}
\icmlauthor{Abbas Abdolmaleki}{goo}
\icmlauthor{Jackie Kay}{goo}
\icmlauthor{Jonas Degrave}{goo}
\icmlauthor{Julian Schrittwieser}{goo}
\icmlauthor{Yuval Tassa}{goo}
\icmlauthor{Jonas Buchli}{goo}
\icmlauthor{Dan Belov}{goo}
\icmlauthor{Martin Riedmiller}{goo}
\end{icmlauthorlist}

\icmlaffiliation{goo}{DeepMind, London, UK}

\icmlcorrespondingauthor{Jost Tobias Springenberg}{springenberg@google.com}
%\icmlcorrespondingauthor{Nicolas Heess}{heess@google.com}

\icmlkeywords{Machine Learning, ICML}

\vskip 0.3in
]



\printAffiliationsAndNotice{\icmlEqualContribution} % otherwise use the standard text.

\begin{abstract}
We present an algorithm for local, regularized, policy improvement in reinforcement learning (RL) that allows us to formulate model-based and model-free variants in a single framework.
Our algorithm can be interpreted as a natural extension of work on KL-regularized RL and introduces a form of tree search for continuous action spaces. 
We demonstrate that additional computation spent on model-based policy improvement during learning can improve data efficiency, and confirm that model-based policy improvement during action selection can also be beneficial. 
Quantitatively, our algorithm improves data efficiency on several continuous control benchmarks (when a model is learned in parallel), and it provides significant improvements in wall-clock time in high-dimensional domains (when a ground truth model is available).
The unified framework also helps us to better understand the space of model-based and model-free algorithms. In particular, we demonstrate that some benefits attributed to model-based RL can be obtained without a model, simply by utilizing more computation.
\end{abstract}

\section{Introduction}

Stable policy optimization in high-dimensions, and continuous action spaces, can be a challenge even in simulation.
In recent years, a variety of deep RL algorithms have been developed, both for the model-free and model-based setting, that aim to tackle this challenge. In continuous control, recent progress on scalable (distributed) algorithms \cite{schulman2017ppo,song2019vmpo} now allows us to solve problems with high-dimensional observation and action spaces end-to-end, provided adequate computation for simulation and learning is available. At the other end of the spectrum, there exist off-policy algorithms \citep{hafnernfqca,lillicrap2015continuous,heess2015learning,abdolmaleki2018maximum,haarnoja2018soft} that have achieved remarkable data-efficiency, and raise hopes that applications of RL in robotics are within reach. For problems in continuous control, recent results that additionally employ learned models of the environment have promised further  data efficiency gains \cite{byravan2019imagined,hafner2019dream}. Analogously, in domains with discrete actions, the combination of model based search, e.g. Monte Carlo Tree Search (MCTS), with RL has recently been shown to be a powerful approach \citep{silver2017mastering,schrittwieser2019mastering,anthony2017thinking}. 

Compared to a more simplistic policy gradient or actor-critic method all of these model-based techniques have one thing in common: they use additional computation (by performing a search or gradient based optimization) either at the time of action selection or in the policy improvement step.
From this observation three questions naturally arise: 1) If data-efficiency can be improved via additional computation how should this computation be spent (i.e. can an off-policy algorithm be as efficient as a model-based algorithm by performing more policy updates)? 
2) If a model of the environment is available, can we conclude that additional search during action selection is beneficial? 
3) Can model-based search and RL be cast into a consistent framework applicable to domains with continuous actions?

In this paper we make an attempt to understand these questions better. In particular we aim to understand how data efficiency and scalability of algorithms for continuous control can be influenced through the use of additional compute during acting or learning. We build on a class of KL-regularized policy iteration schemes \citep{rawlik12,abdolmaleki2018maximum} that separate acting, policy improvement, and learning and thus allow us to flexibly employ parametric policies, value functions, exact or approximate environment models, and search based methods in combination. In this framework we instantiate variants that differ in their reliance on learned approximations (e.g. for Q-function / policy / model) and how they use compute. Among these, our framework allows us to instantiate an off-policy model-free variant, that recovers the MPO algorithm \citep{abdolmaleki2018maximum}, as well as a new algorithm dubbed \ourmethod. This approach can utilize additional compute by searching with an (potentially approximate) environment model -- in a way that is akin to MCTS for decision making in discrete domains. We find that it works robustly with exact and approximate environment models even in high-dimensions.

Using our approach we make a number of observations regarding the questions 1)-3) posed above: \\
\textbf{1.} The literature on model-free algorithms in continuous control has underestimated their data efficiency. Through additional updates of the policy and value-function (additional compute), we can achieve significant improvements. \\
\textbf{2.} At the expense of additional compute a learned predictive model (of rewards and values) can be used during learning for model-based policy optimization; providing a stronger policy improvement operator than a model-free algorithm. \\
\textbf{3.} The approximate model can also be used for local policy improvement at action selection time. This leads to better decision making which consequently improves the data that is collected. Compared to model use during learning, any model bias does not directly affect the parametric policy and the approach provides a further advantage in data efficiency. \\
\textbf{4.} Finally, when an accurate environment model is available, we find that local policy improvement during acting can greatly improve learning speed (in wall-clock time) compared to state-of-the art distributed RL, thereby giving a positive answer to Question 2.

\section{KL regularized Reinforcement Learning}
We consider Markov Decision Processes in which we aim to learn an optimal policy $\pp^*(a | s)$ of actions $a$ in states $s \in \mathcal{S}$. Policies are judged based on a scalar reward signal given in each state $r_t = r(s_t)$ which is discounted over time using a discount factor $\gamma \in [0, 1)$.
Let $\tau = \lbrace (s_0, a_0), \dots (s_T, a_T) \rbrace$ denote a trajectory of length $T$ consisting of state-action pairs where $s_{t+1} \sim f(s_t, a_t)$ is the transition function.
Let $\pp(a | s; \theta)$ be a parameterized (neural network) policy with parameters $\theta$ and $\rho_\pi$ denote the associated distribution of trajectories, when following $\pi$.
Further let $\qp^{1:N} = \lbrace \qp^1, \dots, \qp^N \rbrace$ be a set of $n$ time-varying local
policies such that $\lbrace \pp(a | s_t) > 0 \rbrace \implies \lbrace\qp(a | s_t) > 0\rbrace$. We will also make use of the Kullback-Leibler divergence at state $s_t$ which we denote with $\KL_t = \KL[\qp(\cdot | s_t) \| \pp(\cdot | s_t)])$.

\subsection{The KL regularized objective}

The starting point for our algorithm is the KL-regularized expected reward objective
\begin{equation}
\objKL(\mu, \pi) = \EE_{\tau \sim \rho_\mu} \Big[ \sum_{t} \gamma^{t-1} (r_t - \alpha \KL[\qp(\cdot | s_t) \| \pp(\cdot | s_t)]) \Big], \label{eq:objectiveJ}
\end{equation}
with associated regularized action-value function $Q^\mu_\pi(s, a) = r_t + \gamma  \EE_{s_{t+1}}[V^\mu_\pi(s_{t+1})]$ and value function $V^\mu_\pi(s) =  \EE_{\rho_\mu} [\sum^{|\tau|}_{t=1} \gamma^{t-1} (r_t - \alpha \KL_t) | s_0 = s]$.
This objective can be motivated from different perspectives. It arises naturally in the control as inference framework \citep{toussaint2006probabilistic,rawlik12}, and is closely related to the objective of maximum entropy RL where the relative entropy (KL) is replaced by a simple entropy term \citep{haarnoja2017reinforcement,schulman2017equivalence}. Broadly speaking, existing algorithms optimizing this objective can be grouped into two classes: those that optimize $\objKL$ only with respect to $\qp$, and those that optimize $\objKL$ with respect to both $\qp$ and $\pp$; typically in an alternating fashion. While the former solve a regularized objective in which expected reward and closeness to the ``reference distribution'' $\pi$ are traded-off against each other, the latter can converge to the optimum of the expected reward objective \citep[e.g.][]{rawlik12,abdolmaleki2018maximum} and are also referred to as EM policy search algorithms \citep{Deisenroth2013}.

Here, we consider the alternating optimization of $\mathcal{J}$ wrt. $\qp$ and $\pp$. The optimization occurs iteratively, repeating the following two steps (referred to as E-and M-step respectively, in analogy to the EM algorithm for statistical inference):
\textbf{E-step} Optimization of $\qpi = \arg\max_\mu \mathcal{J}(\mu, \ppi)$ given a fixed $\ppi$.
\textbf{M-step}
Optimization of $\mathcal{J}$ with respect to $\pi$ given $\mu^{(i)}$; which amounts to minimizing the average $\KL$.

\section{KL regularized K-step Improvement}
We make several observations about $\objKL$ -- from which we will then derive a model based policy optimization scheme:

\textbf{A relation to the RL objective.} When $\pi = \mu$ then
$
\objKL(\pi, \pi) = \EE_{\tau \sim \rho_\pi} \left [ \sum_t \gamma^{t} r_t \right ], 
$
since the KL term vanishes. 

\textbf{Optimality conditions from existing work.}
Maximizing $\objKL$ with respect to $\mu$ gives the optimal policy $\mu^*(a|s) \propto \pi(a|s) \exp( \nicefrac{Q^{\mu^*}_{\ppi}(s,a)}{\alpha})$ \citep[see e.g.][]{rawlik12,fox2015taming}.
We thus have $\objKL(\mu^*, \pi) \geq \objKL(\mu, \pi)~\forall \mu$.  \\
It turns out that the following is also true: for any $\pi$ and $\mu_{1}(a|s) \propto \pi(a|s) \exp (\nicefrac{Q^{\pi}_\pi(s,a)}{\alpha})$ we have  $\objKL(\mu_1, \pi) \geq \objKL(\pi, \pi)$ \citep[see e.g.][]{abdolmaleki2018maximum}. We can think of $\mu_1$ as picking the action that is soft-optimal under the assumption that at the next time step we resort to acting according to the current policy $\pi$. $\mu_1$ thus amounts to a local improvement over $\pi$, and we refer to it below as one-step improved.
This insight is exploited e.g.\ in the algorithm of \citet{abdolmaleki2018maximum}, which iterates three steps (1) estimating $Q^\pi_\pi$, (2) optimizing for $\mu_1$, and (3) moving $\pi$ towards $\mu_1$; corresponding to a policy iteration scheme that repeatedly performs one-step improvements of the policy.

\textbf{From one-step to K-step policy improvement.}
Going beyond one-step improvements we can consider a sequence of
$K$ local policies $\mu_{1:K} = \lbrace \mu_{K}(\cdot | s_1), \dots,  \mu_{1}(\cdot | s_k)\rbrace$ that are individually optimized based on the result of the following nested K-step optimization adapted from Equation \eqref{eq:objectiveJ}:
\begin{equation}
\begin{aligned}
    &\mu^{*}_{1:K}(a |s) = \arg \max_{\mu_{1:K}} \objKL^K(\mu_{1:K}, \pi) \\ = &\arg \max_{\mu_{1:K}} \EE_{\mu_{1:K}} \Big [ \gamma^{K} V^\pi_\pi(s_{K+1}) + \sum_{t=1}^{K} \gamma^{t - 1} (r_{t} - \alpha KL_{t}) \Big]
\end{aligned}
\end{equation}
where $s_1 = s, a_1 = a,$ and using the short-hand notation $r_t = r(s_t, a_t)$, $\KL_t = \KL[ \qp(\cdot|s_t) \| \pp(\cdot | s_t)]$.
The solution to this optimization corresponds to the policy that acts soft-optimally for the next $K$ steps, and subsequently resorts to acting according to $\pi$; and it bears resemblance to the $K$-step greedy policy defined in \cite{efroni2018combine} albeit for a regularized objective.
Here, too, we find that
$
    \objKL(\mu^*_{1:K}, \pi) \geq \objKL(\pi, \pi)~~~\forall \pi,
$
since each local policy in $\mu^*_{1:K}$ improves on $\pi$ by at least as much as the one-step improvement.

\begin{observation} 
We can express $\mu^*_{1:K}$ recursively via a K-step soft-optimal Q-function (see supplementary for details):
\begin{equation}
\begin{aligned}
    \mu^{*}_{K}(a|s) &\propto \pi(a|s) \exp \Big( Q_{\pi}^{*_{K-1}}(s,a) / \alpha \Big) \\
    Q_{\pi}^{*_k}(s,a) &= \Bigg\{\begin{array}{ll}
    r(s) + \gamma \EE_{s' \sim p_{\mu^*_{k}}}[ V^{*_{k-1}}_{\pi}(s') ] &\text{if } k > 1 \\
    r(s) + \gamma \EE_{s' \sim p_{\mu^*_{k}}}[ V^{\pi}_{\pi}(s') ]  &\text{else}
    \end{array} \\
    V^{*_k}_{\pi}(s) &= 
    \alpha \log \int \pi(a|s_t) \exp \Big( Q_{\pi}^{*_k}(s,a) / \alpha \Big) d a. 
\end{aligned}
\label{eq:QMuPiKstep}
\end{equation}
\end{observation}
In addition, if all $\lbrace Q_{\pi}^{*_k} \rbrace_{k=1}^K$ are given then it is also easy to sample approximately from an estimate $\hat{\mu}_K \approx \mu^*_{K}$, e.g.\ via the following self-normalized importance sampling scheme:
\begin{equation}
    \begin{aligned}
        a \sim \hat{\mu}_K(\cdot | s_t) &=  \text{Categorical}(a^{1:M}; \nicefrac{w^{1:M}_K}{ \sum_{j=1}^M w^j_K)}, \\
        \text{with } w^{1:M}_K &= \lbrace \exp(Q^{*_{K}}_\pi(s, a^j)/\alpha) \rbrace_{j=1}^M, \\
        a^{1:M} &= \lbrace a^1, \dots, a^M \rbrace \sim \pi(\cdot | s).
    \end{aligned}
    \label{eq:resample}
\end{equation}
Thus, to sample from $\mu^*_K( \cdot | s)$ we simply sample $M$ actions from $\pi(\cdot|s)$ and importance weight with the exponentiated soft-Q values $Q^{*_{K}}_\pi(s, a)$, which is possible both for discrete and continuous actions without discretization.

A number of different schemes can be used to estimate $\mu_{1:K}$. If $\mu$ is parametric we can directly follow the gradient of $\objKL^K(\mu_{1:K}, \pi)$ with respect to $\mu_{1:K}$ resulting in a K-step KL regularized policy gradient \citep[related to e.g][]{schulman2017equivalence,haarnoja2018soft,Bhardwaj20arxiv}, as described in more detail in the supplementary. 
Alternatively we can attempt to directly estimate $Q^{*_{K}}_\pi$ using the recursive form of Equation \eqref{eq:QMuPiKstep}, from which $\mu^*_{K}$ follows immediately as above.
This choice allows us to instantiate a family of policy iteration schemes in which we alternate between estimating $\mu_{1:K}^*$ for the current $\pi$ and updating $\pi$ towards $\mu_K^*$:
\textbf{i) perform an E-Step} by locally estimating $\mu^{(i)}_{1:K} \approx \mu^*_{1:K}$ for reference policy $\pi = \pi^{(i)}$ according to the recursion in Eq. \eqref{eq:QMuPiKstep}; \textbf{ii) perform an M-Step} by minimizing $\min_\pi \objKL^K(\mu^{(i)}_{K}, \pi)$ which amounts to fitting $\pi^{(i+1)}$ to samples from $\mu^{(i)}_{K}$: 
\begin{align}
\ppnext = \arg \min_{\pi} \EE_{s\sim \tau_\mu} [ \KL [ \mu^{(i)}_{K}(\cdot | s) || \pi(\cdot | s) ]],
\end{align}
and we fit a parametric approximation to $Q^\ppnext_\ppnext$ for bootstrapping in the next E-step.

For such a policy iteration scheme to be effective, a lightweight estimate of $\mu_K^*$ is desirable. In a model free setting this is largely impractical for $K>1$ since a Monte-Carlo estimator of Eq. \eqref{eq:QMuPiKstep} may be very hard to construct (it would at least require priviliged access to the environment; i.e. the ability to reset to any given state to perform the required K-step rollouts). For $K=1$ we recover the method from  \citet{abdolmaleki2018maximum} which uses a parametric $Q^\pi_\pi$ to construct a Monte Carlo estimate of $\mu_1$ in the E-step.

In the case where a model of the environment is available, or can be learned, sample based estimation of $Q^{*_K}_\pi$ from Eq. \eqref{eq:QMuPiKstep} becomes practical -- as we will show below. The only caveat in this case is that the naive approach would require the evaluation of a full tree of trajectories (up to depth K).

\section{Policy improvement with \ourmethod}
\label{sec:LocalSearch}
In this section we propose an algorithm for approximately sampling from $\mu_K^*$ that removes the, impractical, requirement of building a full tree up to depth K.

We assume deterministic system dynamics and the availability of a model $s_{t+1} = f(s_{t}, a_{t})$. In this case we can obtain a particle approximation to $\mu_K^*$ using a form of self-normalized importance sampling that bears similarity to MCTS. 
We first describe the procedure in its exact form and then a practical, approximate algorithm. 

\textbf{Unbiased solution.} For brevity of notation, and without loss of generality, let $\alpha = 1$. For deterministic systems
we can estimate $\exp Q^{*_{K}}_\pi(s, a)$ by building a tree, of depth $K$ in the following way: starting from $s$ we recursively sample $M$ actions per state according to $\pi$, i.e.\ $a_j \sim \pi(\cdot | s)$, up to a depth of $K$ steps. Arrows are labeled with actions. Leaves are set to $Q^\pi_\pi(s,a)$. We can then compute an approximation to $\exp Q^{*_{K}}_\pi(s, a)$ recursively using the approximate value function $\hat{V}^{d+1}(s) = \log \frac{1}{M} \sum_{j=1}^M \exp( r(f(s,a_j)) + \gamma V(f(s,a_j))$ starting from the leaves $\hat{V}^1(s) = \frac{1}{M} \sum_{j=1}^M \exp( Q^\pi_\pi(s,a_j) )$. Then $\hat{Q}^K(s,a) = r(s) + \gamma \hat{V}^{K-1}(f(s,a))$. It is easy to see that $\EE [ \exp \hat{Q}^K(s,a)] = \exp Q^{*_{K}}_\pi(s, a)$, i.e.\ $\exp \hat{Q}^K(s,a)$ is an unbiased estimate of $\exp Q^{*_{K}}_\pi(s, a)$. We can then sample from $\hat{\mu}_K$ as described in eq.\ (\ref{eq:resample}) using $\hat{Q}^K$ instead of $Q^{*_{K}}_\pi$.

\textbf{Practical algorithm.} Building the full tree of depth $K$, with branching factor $M$, would be computationally expensive. Instead, we build the tree greedily, following the procedure in Algorithm \ref{alg:treepi}. Rather than expanding all $M$ actions it samples from the current approximation to $\mu_K$. Intuitively, the procedure consists of two interleaved steps: 1) forward rollouts starting in $s_t$ using the model $f$ and our current best guess $\hat{\mu}_{1:K}$ for the locally improved policy, 2) soft-value backups to update our current best estimate of $\hat{Q}^K[s_t, a] \approx Q^{\mu_K}_{\pp}(s_t, a)$ -- which in turn results in a better estimate $\hat{\mu}_{1:K}$. These steps are repeated until $N$ total rollouts have been performed. We note that, the algorithm only requires sampling from the current policy $\pi$ and thus works for both continuous and discrete action spaces. Although for $N<M^K$ we cannot guarantee that the MC estimate $w^j = \exp \hat{Q}^K(s,a^j)$ of $\exp Q^{*_{K}}_\pi(s, a)$ (with $j=1\dots M$) is unbiased, it is easy to ensure that it will be after $N=M^K$ rollouts (by preventing fully expanded subtrees from being revisited).
We note that this analysis relies on the environment being deterministic, in the stochastic case the log-sum-exp calculation introduces an optimism bias. 

\begin{algorithm}[tb]
   \caption{Tree Policy Improvement (TreePI)}
   \label{alg:treepi}
\begin{algorithmic}
   \STATE {\bfseries Input:} state $s_t$, policy $\pi^{(i)}$, approximate value function $\hat{Q}^\ppi_\ppi$, branching factor $M$, number of rollouts $N$, maximum depth $K$, multiplier $\alpha$
   \STATE {\bfseries Output:} Importance weights $w^{1:M}_K$ and corresponding samples $a^{1:M}_t$ such that $\sum_{j=1}^M w^j_k \log \pi_\theta(a^j | s_t) \approx \EE_{a \sim \mu^{(i)}_K(\cdot | s_t)} [\log \pi_\theta(a^j | s_t)]$
   \STATE {\bfseries Initialize:} $\mathcal{T}[d] = \lbrace \rbrace \ \forall: d < K$
   \FOR{$i=1$ {\bfseries to} $N$}
   \STATE $s_\text{prev} = \text{None}; a = \text{None}; s = s_t; d = 0$
   \STATE // forward rollout
   \WHILE{$s \in \mathcal{T}[d]$, $d < K$}
   \STATE // fetch transition from tree
   \STATE (\_, \_, $a^{1:M}) = \mathcal{T}[d][s]$
   \STATE // resample actions according to Eq. \eqref{eq:resample}
   \STATE $w^{1:M}_d = \lbrace \exp(\hat{Q}^{d}[s, a^j] / \alpha) \rbrace_{j=1}^M$
   \STATE $a \sim \text{Categorical}(a^{1:M}, \text{prob}= w^{1:M}_d/\sum_j w^j_d)$
   \STATE // remember state and perform transition
   \STATE $s_\text{prev} = s; d = d + 1$
   \STATE $s = f(s, a)$
   \ENDWHILE
   \STATE // Insert new transition into tree
   \STATE $T[d][s] = \text{Node}(s_\text{prev}, a, \lbrace a^1_t, \dots, a^M_t \rbrace \sim \pp^{(i)}(\cdot | s))$
   \STATE $\hat{Q}^d[s_t, a^{1:M}_d] = \lbrace \hat{Q}^{\pi}_{\pi}(s_t, a^1_d), \dots,  \hat{Q}^{\pi}_{\pi}(s_t, a^M_d) \rbrace$
   \STATE // soft-backup
   \WHILE{$s$ {\bfseries not} \text{None}}
   \STATE $(s_\text{prev}, a, a^{1:M})= \mathcal{T}[d][s]$
   \STATE $V  = \alpha \log \frac{1}{M}\sum_{j=1}^M \exp(Q^d[f(s, a^j), a^j] / \alpha)$
   \STATE $\hat{Q}^{d+1}[s_\text{prev}, a] = r(s) + \gamma V$
   \STATE $d = d - 1$; $s = s_\text{prev}$ 
   \ENDWHILE
   \ENDFOR
   \RETURN $a^{1:M}_t$, $w_0^{1:M}$
\end{algorithmic}
\end{algorithm}

\section{Learning parametric estimators}
\label{sec:ModelLearning}
In the general RL setting it is not always reasonable to assume knowledge of the environment model $f$. 
For the K-step policy improvement operator from Section \ref{sec:LocalSearch} to be applicable in these settings a model has to be learned jointly with a parametric policy and Q-function estimator.
In these cases, and to allow for seamless transition from the model-free to model-based setting, we learn a predictive model of future rewards and Q-values. This bears some similarity to recent work \citep{schrittwieser2019mastering}; although we here learn Q-predictions in an off-policy manner. We focus on assessing the difference between search-based policy improvement and model-free learning only. We hence do not aim to model future observations to prevent corroboration of our experimental results with issues due to modelling high-dimensional physical systems (e.g. predicting the dynamics of a humanoid walking). On the other hand, it is possible that in some domains learning an observation model could lead to further improvement; both for the model-free and search based instances, as argued in some recent works \citep{alexlee2019,byravan2019imagined,hafner2019planet}. 
\begin{figure*}[t]
    \centering
    \includegraphics[width=\textwidth]{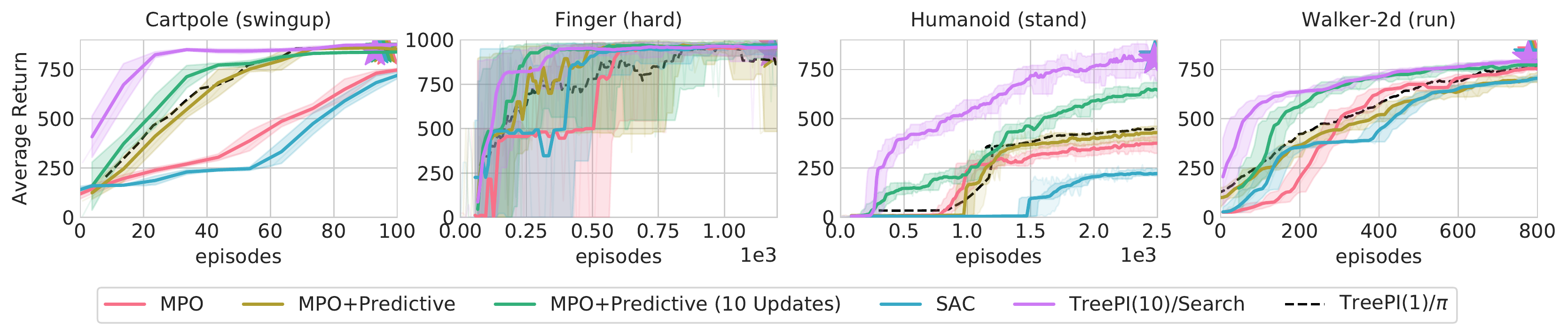}
    \vspace{-0.5cm}
    \caption{Comparison of TreePI (with depth K=10 and executing the search policy during at decision time) against state-of-the-art algorithms on control suite domains. Search improves slightly over MPO with 10 updates per step and a predictive model. MPO without predictive model loss -- as well as SAC -- is less data efficient. We plot the median performance (over 5 runs). All algorithms reach similar performance at convergence (overlapping stars denote median after 20k episodes).}
    \label{fig:suite_comparison}
\end{figure*}

For training, we assume access to a replay buffer containing trajectory snippets, $\mathcal{B} = \lbrace \tau^{1:T}_1, \dots, \tau^{1:T}_{|\mathcal{B}|} \rbrace$, of observations and actions, $\tau_i = \lbrace(o_1, a_1, r_1), \dots, (o_T, a_T, r_T)\rbrace$. We denote by $s_t f_{\phi_\text{enc}}(o_t)$ a parameterized encoding function, mapping from observations to a ``latent'' state and by $f_{\phi_\text{trans}} = (s_t, a_t)$ a learned transition model. Let $s_t = f_\phi(o_1, a_{1:t-1}) = f_{\phi_\text{trans}}(f_{\phi_\text{enc}}(o_1), a_{1:t-1})$ be the prediction of the state at time $t$ after encoding observation $o_1$ and applying the approximate transition model, conditioned on observed actions, for $t$ time-steps. 
We train a predictive model of rewards (loss $\mathcal{L}_r$), Q-values (with loss $\mathcal{L}_Q$) and actions (with policy loss $\mathcal{L}_{\pi_\theta}$). This amounts to finding the minimum of the combined loss:
\begin{equation}
\begin{aligned}
    \mathcal{L} & = \EE_{\tau \sim \mathcal{B}} \left[ \sum_{t=1}^T \mathcal{L}_r + \mathcal{L}_Q + \mathcal{L}_{\pi_\theta}  \Big|  s_t = f_{\phi}(o_1, a_{1:t-1}) \right], \\
    \mathcal{L}_r &= \left( r_t - r_{\phi_r}(s_t) \right)^2, \\
    \mathcal{L}_Q &= \Big( r_t + \gamma \mathbb{E}_{\pi^{(i)}}[\hat{Q}^{\pi_\theta}_{\pi_\theta}(s_t, \cdot; \phi_Q')] - \hat{Q}^{\pi_\theta}_{\pi_\theta}(s_t, a_t; \phi_Q) \Big)^2, \\
    \mathcal{L}_{\pi_\theta} &= \KL[\hat{\mu}^K(\cdot | s_t) \| \pi_\theta(\cdot | s_t)] + R(\pi_\theta, \ppi, s_t) \\ &\approx \sum_{j=1}^M - \log \pi_\theta(a^i_t | s_t) w^i_t + R(\pi_\theta, \ppi, s_t) \\
    \text{where }& a^{1:M}_t, w^{1:M}_t = \text{TreePI}(s_t, \pi^{(i)}, \hat{Q}^{\pi^{(i)}}_{\pi^{(i)}}, M, N, K, \alpha),
\end{aligned}
\label{eq:modelLosses}
\end{equation}
with $\phi_r, \phi_Q, \theta$ in the above denoting the parameters of the learned reward predictor, Q-value function and policy respectively. Note that the above slightly abuses notation, using $s_t$ to refer to the latent states, to simplify the presentation; and we assume $s_1 = f_\text{enc}(o_1)$. We use a target network (with parameters $\phi'_Q$ that are copied every periodically from $\phi_Q$) to stabilize the TD loss $\mathcal{L}_Q$.
Further, $R$ denotes an additional regularization term that prevents over-fitting of the policy $\pi_\theta$ to the M samples. This can be crucial for avoiding collapse of the policy distribution in the low sample and small temperature regime (i.e. small $M = 20$, small $\alpha$) that we typically find ourselves in. While the form of regularization can be chosen freely, recent work in RL has found KL regularization via a
trust-region constraint to be particularly effective \citep{schulman2015trust,abdolmaleki2018maximum}. We thus let $R(\pi_\theta, \ppi, s_t) = \eta (KL[\pi^{(i)}(\cdot | s_t) \| \pi_\theta(\cdot | s_t)] - \epsilon_\text{KL})$, set $\epsilon_\text{KL} = 0.005$ to a small value and optimize for the Lagrangian multiplier $\eta$ together with other parameters via gradient descent / ascent, solving for:
$\arg \max_\eta \min_{\phi,\theta} L$. 
Finally, after $500$ steps we set $\pi^{(i+1)} = \pi_\theta$ and start the next round of optimization -- i.e. we perform partial optimization for computational efficiency. A listing of the procedure is given in Algorithm 1 in the supplementary material.

\section{Experiments}
To understand when and how TreePI can improve performance we
consider several experimental scenarios. We start with experiments on the DeepMind control suite \citep{deepmindcontrolsuite2018} which explore the utility of our algorithm in combination with a learned model as described in Section \ref{sec:ModelLearning}. We focus on data efficiency and compare to several strong off-policy baselines for learning from scratch.

In a second set of experiments, we attempt to disentangle the effects of search from the problem of learning predictive models. We assume access to the true model $f$ of the environment; inserting true states in the calculation of Eq. \ref{eq:modelLosses} and replacing the learned reward $r_{\phi_r}$ with the true reward function $r$. The focus here is on the best use of a fixed compute budget to achieve fast learning.

\subsection{Control Suite with a Learned Predictive Model }
\label{sec:Experiments:DataEfficient}

\paragraph{Experimental Setup}
We experiment with four domains from the DeepMind control suite: i) the 'Cartpole (swingup)' task, where the control is one-dimensional and the goal is to swing-up and balance a pole attached to a cart, ii) the 'Finger (hard)' task where the goal is to spin a rotating object to a target orientation with a two-joint robot, iii) the 'Humanoid (stand)' task where a humanoid with 21 degrees of freedom should stand up from a randomly initialized position, iv) The 'Walker-2d (run)' task where a simpler, two-dimensional, body should run as fast as possible. 

\paragraph{Model setup and Baselines}
We use feed-forward neural networks to represent the different components $f_{\phi_\text{enc}}$, $f_{\phi_\text{trans}}$, $r_{\phi_r}$, $\hat{Q}_{\phi_Q}$, $\pi_\theta$ in \textbf{\ourmethod}. Full details on the network setup and hyperparameters are given in the supplementary material. The reward, Q and policy networks $r_{\phi_r}, \hat{Q}_{\phi_Q}, \pi_\theta$ operate on the shared representation obtained from $f_{\phi_\text{enc}}$ and  $f_{\phi_\text{trans}}$; and we use a network predicting mean and variance of a Gaussian distribution as the policy $\pi_\theta$.
The hyperparameters for all methods were tuned based on initial experiments on the Walker-2D domain and then fixed for the remaining experiments. If not otherwise noted we set $M=20$, $K=10$ and $N=200$ in the experiments for \ourmethod. We use two versions of \ourmethod: (a) we execute the current policy $\pi$ at action selection time (and hence only perform search when updating the policy) denoted by \textbf{TreePI/$\pi$}; and (b) at action selection time, we perform an additional search to draw a sample from $\hat{\mu}_K$ for the current state (cf. Eq \eqref{eq:resample}), denoted by \textbf{TreePI/search}. 

We consider two sets of off-policy RL baselines to disentangle improvements due to the network (and additional predictive losses) from improvements due to the policy optimization procedure itself. First, we compare to two state-of-the-art RL algorithms: \textbf{MPO} \citep{abdolmaleki2018maximum} and \textbf{SAC} \citep{haarnoja2018soft}. We align their network architecture with the \ourmethod setup to the extent possible (using separate policy and Q-function networks; see the supplementary) and run them using the same software infrastructure to minimize differences due to tooling.
Second, we compare to a variant of MPO that performs standard model-free learning but uses the same network architecture and additional predictive losses for reward, Q-function and policy as \ourmethod.
This baseline (\textbf{MPO+Predictive})
uses the model merely to provide auxiliary losses and not for policy improvement.

\paragraph{Results}
The main results are depicted in Figure \ref{fig:suite_comparison}. 
First, we observe that simply adding a predictive model of rewards and values to a standard off-policy learner (\textbf{MPO+Predictive}) results in improved data-efficiency in the Cartpole and Finger domain when compared to well tuned versions of state-of-the-art agents (\textbf{SAC}, \textbf{MPO}).
Second, \ourmethod with $K=10$ (\textbf{TreePI(10)/Search}) results in improved data-efficiency in three of the four domains; remarkably achieving near optimal performance in all domains in less than 2000 episodes. 

\begin{figure}
    \centering
    \includegraphics[width=\columnwidth]{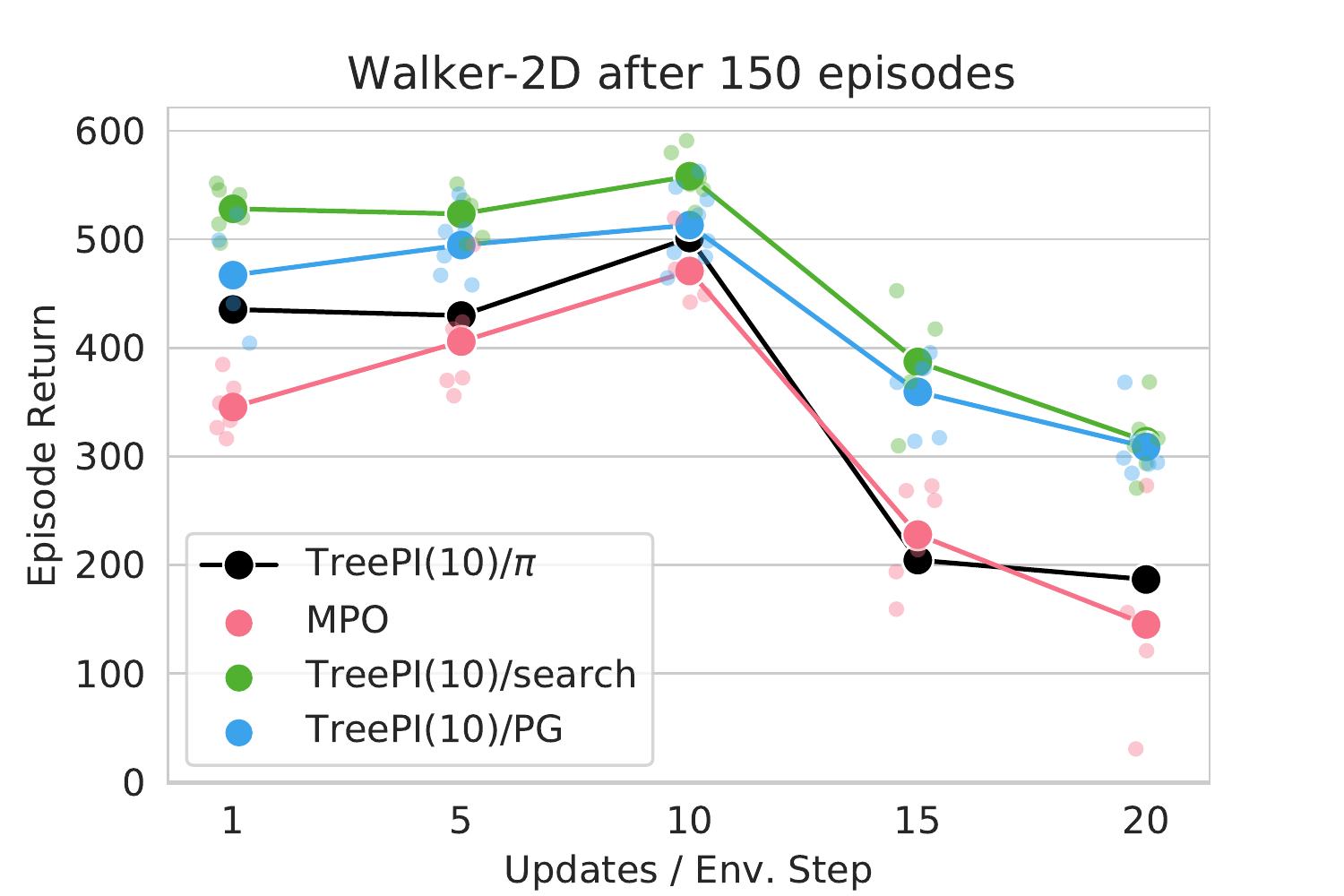}
    \vspace{-0.5cm}
    \caption{Comparison of different compute budgets on the learner and actor side for the Walker-2D measured after 150 episodes. Performing more learning steps per collected experience improves performance (i.e. data-efficiency) until overfitting occurs.}
    \vspace{-0.5cm}
    \label{fig:search_in_depth}
\end{figure}

To analyze this improvement further, we test how much \ourmethod gains from the model-based search and how much from the fact that it uses additional compute (via model rollouts). To isolate these effects, we varied the number of Policy and Q-function updates performed per environment step for different methods. 
Results are presented in Figure \ref{fig:search_in_depth}, where we plot the performance against updates per environment step; displayed at a specific time during training (150 episodes).
In addition we also plot the performance for \textbf{TreePI(10)/$\pi$} -- i.e. using TreePI but executing $\pi$, thus no search during action selection.  
All methods gain in data-efficiency as the number of updates -- and thus the use of compute -- grows up to 10 (see e.g.\  \citealp{popov2017data} for related results). Additional updates result in premature convergence of the policy (overfitting), reducing performance -- and also resulting in sub-optimal behavior at convergence; not shown in the plot. Interestingly, \textbf{MPO+Predictive} gains in data-efficiency at a larger rate with more updates, almost catching up to \textbf{TreePI(10)/$\pi$} at 10 updates per step. To better appreciate this result we also plot \textbf{MPO+Predictive} in Figure \ref{fig:suite_comparison} which obtains performance much closer to \ourmethod. In addition, running \ourmethod with a depth of $K=1$ and executing $\pi$  (\textbf{TreePI(1)/$\pi$} Figure \ref{fig:suite_comparison}) recovers MPO+Predictive as expected (except for differences in setting $\alpha$, see supplementary). Thus, even though some of the improvement can be attributed to TreePI being a better policy improvement operator -- it is consistently better than MPO at lower numbers of updates -- %model-free algorithms may also not be making full use of the data they are collecting. 
spending more compute on policy updates can partially alleviate this difference. %We thus conclude that, 
Some of the data-efficiency gains attributed to model-based methods in the literature may thus be obtained in a model-free setting; a result similar to recent observations for RL in discrete domains \citep{hasselt2019model}. 

Nonetheless, additional search at action selection time (\textbf{TreePI(10)/search}) still results in an improvement over all other algorithms; even at a high number of updates per step. This, to some extent, supports the hypothesis that fast adaptation during action selection combined with 'slower' learning can result in effective policy improvement --  which underlies both the traditional paradigm of model predictive control (MPC; see e.g. \citealp{MPCMac02} for an overview and \citealp{lowrey2018plan} for a modern variant) and the more recent ideas of Expert-Iteration \citep{anthony2017thinking} and AlphaZero / MuZero \citep{silver2017mastering,schrittwieser2019mastering} for discrete search in games.

We speculate that the advantage of additional search at action selection time is that it temporarily shifts the action distribution towards a better policy but does not change the policy parameters. This benefit remains even when more gradient descent steps on the policy parameters would lead to overfitting. The ability to replan after every step given the true environment states also mitigates the effect of model error.
To lend further support to this finding we tested a different mechanism to temporarily change the policy: we paired TreePI with a policy gradient based update, instead of search, at action selection time -- taking 10 gradient steps w.r.t. the policy parameters based on 100 K-step rollouts each. This is similar to a recent exploration on replacing search with policy gradients in discrete domains presented in \citep{pgsearch}. This change results in almost the same improvement (\textbf{TreePI(10)/PG} in Figure \ref{fig:search_in_depth}) over \textbf{TreePI(10)/$\pi$}.
Fully characterizing this phenomenon is an interesting direction for future work.

\subsection{Humanoid Domains with a Known Model}
For the experiments with a known environment model and reward we consider more challenging domains and attempt to answer the question: can \ourmethod help us to allocate computational budget more efficiently to `solve' them faster?

\paragraph{Experimental setup}
We consider three difficult, high-dimensional domains, using humanoid bodies.
First, we perform experiments with the `Humanoid (run)' task from the Control suite (see above). 
Second, we consider the problem of reproducing highly agile motion-capture data with the 
the complex humanoid body of \citet{merel2018hierarchical}.
Deep RL approaches are currently popular for producing controllers that track motion-capture data, but the training process can be quite slow \citep[e.g.][]{peng2018deepmimic}. Earlier 
research has used model-based search to control humanoids \citep{tassa2012synthesis, hamalainen2014online} and match reference movements \citep{liu2010sampling}.
Our approach is able to interpolate between model-free deep RL and sample-based planning. 
Concretely, we evaluate a setup similar to \citet{peng2018deepmimic, merel2018hierarchical} 
and train a policy to reproduce a cartwheel and backflip motion from the CMU Motion Capture Database\footnote{http://mocap.cs.cmu.edu/} (see the supplement for a full description of the task). We note that this task is non-trivial since the simulation is physically accurate and the humanoid's actuation, weight and size are different from the people that executed the recorded motion (in fact, exact replication of the reference motion may not be possible).

We use a fast, distributed, implementation of \ourmethod (with the search written in C++) and compare to high-performance distributed model-free RL algorithms. We use a distributed actor-learner setup analogous to \citet{espeholt18a}. To keep the comparison as fair as possible, we restrict \ourmethod to 16 asynchronous actors using 32 threads for expanding the tree. We use up to 6000 actors for the RL algorithms, at which point they send 100x more data back to the learner and perform 10-50x more total environment interactions than \ourmethod (counting all transitions within the search) at 10x the compute cost. We set the branching factor to $M=50$, and the depth to $K=10$ and experiment with varying $N$ for TreePI, $\alpha = 0.1$ was used throughout. We use both an off-policy algorithm \textbf{MPO} as well as an on-policy algorithm \textbf{PPO} \citep{schulman2017ppo} as baselines. We keep network architectures as similar as possible. 
Full details of the experimental setup are given in the supplemental material. 
For TreePI search on the learner could become a computational bottleneck. Fortunately, with a larger number of actors that all perform search to choose high-quality actions, this data is of sufficient quality so that we can simply maintain a small replay buffer (containing 100k time-steps) and train the policy to match actors action choices directly.
We thus adjust policy learning to simply maximize the likelihood of actions sampled from this buffer. That is, we change the policy loss in Equation \eqref{eq:modelLosses} to $L_{\pi_\theta} = \log \pi(a_t | s_t) + R(\pi_\theta, \pi^{(i)}, s_t)$ for $a_t, s_t \in \tau$. The rest of the losses are kept unchanged.

\begin{figure}
\centering
\includegraphics[width=.98\columnwidth]{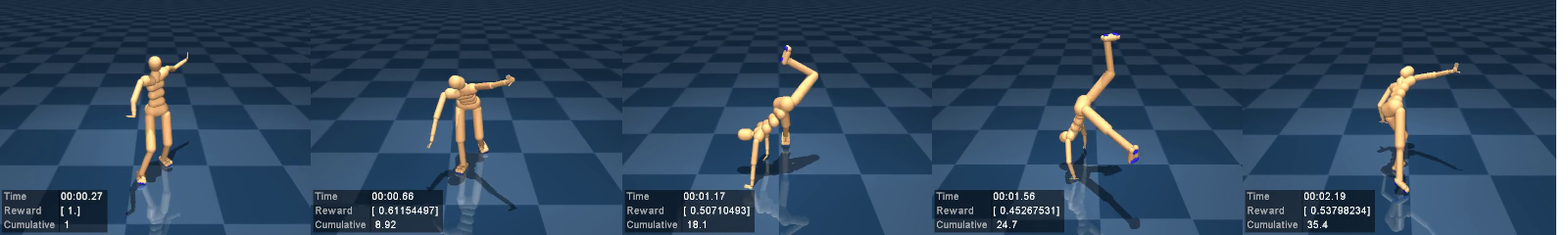} \\
\ \includegraphics[width=.98\columnwidth]{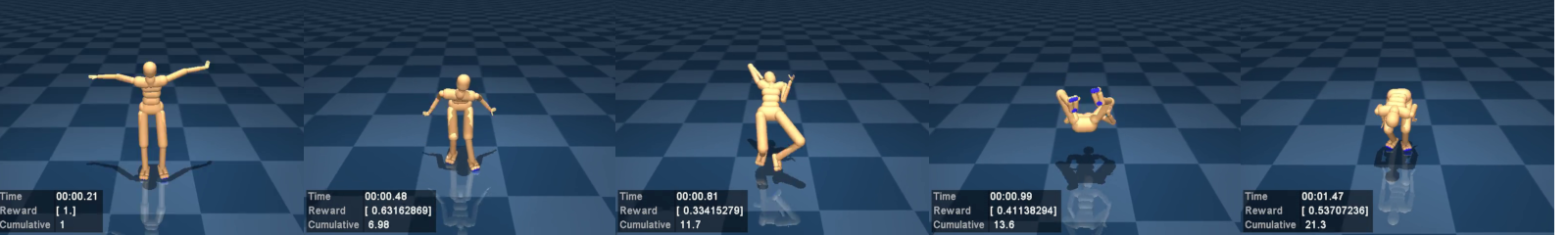}
\label{fig:cartwheel}
\caption{Visualizations of \ourmethod performing the Cartwheel and Backflip motion-capture tasks. A full \emph{video of the respective trajectories can be found in the supplementary material}.}
\vspace{-0.5cm}
\end{figure}

\begin{figure*}[t]
    \begin{minipage}[t]{0.6\linewidth}
    \vspace{0pt}
    \centering
    \begin{tabular}{c|c|c|c}
         Algorithm / Task & Humanoid (run) & Cartwheel & Backflip 
         \\
         & 3h / 48h & 6h / 12h / 48h & 6h / 12h / 48h \\
         \hline
         TreePI (2k roll.) & \textbf{786.5/840.7} & \textbf{48.7 / 61.2 / 61.2} & \textbf{40.2 / 54.6 / 58.9} \\
         TreePI (500 roll.) & 756.8/\textbf{839.7} & 32.9/49.2/\textbf{60.5} & 29.7/42.8/\textbf{58.2}\\
         TreePI (10 roll.) & 374.7/\textbf{843.5} & 13.8/30.3/\textbf{58.7} & 11.2/29.6/\textbf{56.4}\\
         MPO (64 actors) & 374.6/\textbf{837.3} & 7.8/15.3/54.3 & 9.1/18.7/49.7 \\
         MPO (6k actors) & 324.9/\textbf{842.8} & 9.4/23.6/\textbf{60.3} & 12.6/27.4/\textbf{57.7} \\
         PPO (6k actors) & 526.3/\textbf{839.2} & 6.3/28.4/\textbf{59.9} & 14.7/29.1/\textbf{58.4}
    \end{tabular}
    \end{minipage}%
    \begin{minipage}[t]{0.4\linewidth}
    \vspace{0pt}
    \centering
    \ \ 
    \includegraphics[width=0.74\columnwidth]{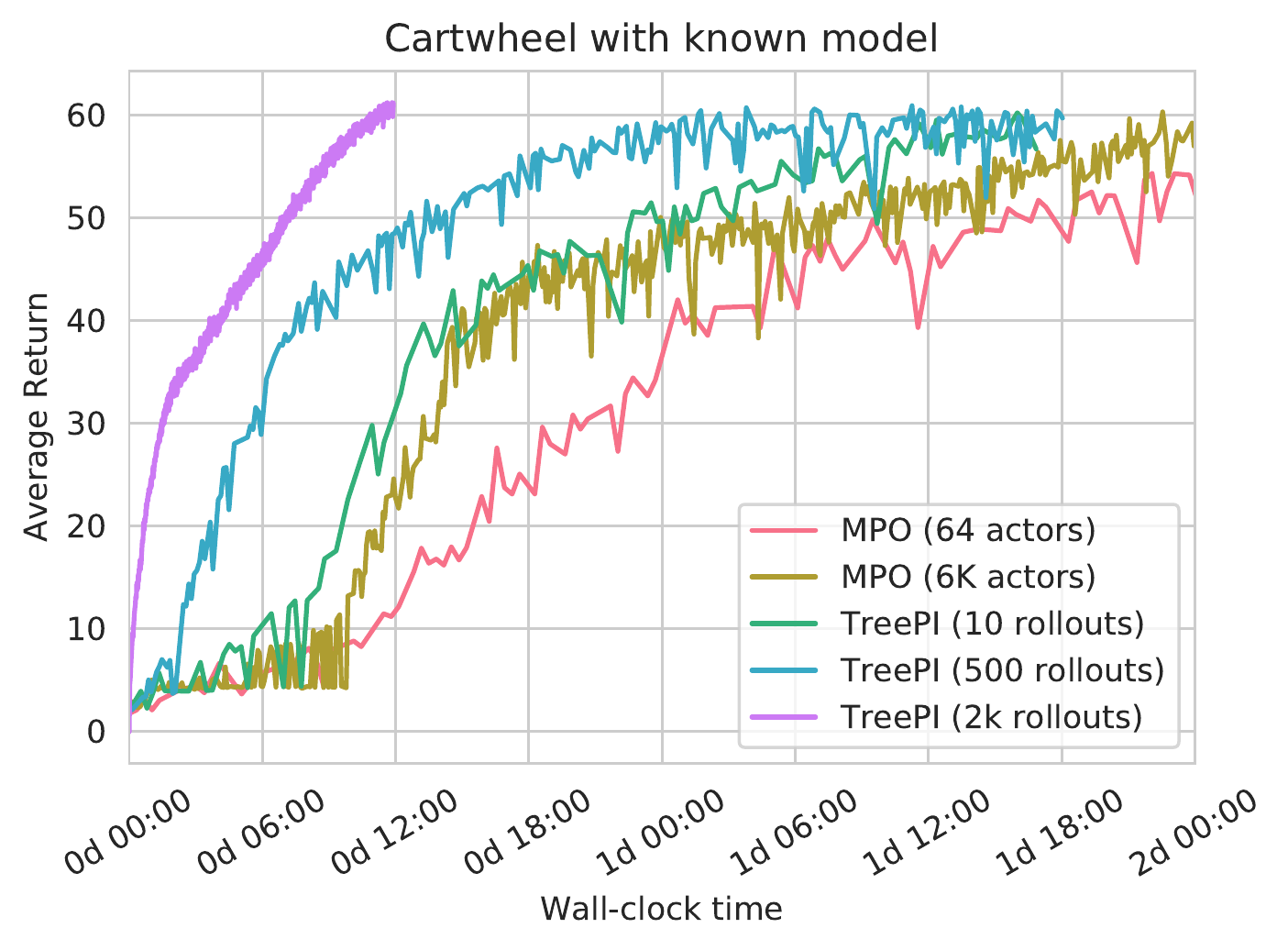}
    \end{minipage}
    \caption{Left: Performance wrt. wall-clock time for solving the three high-dimensional humanoid control tasks (average over 20 test episodes). \ourmethod results in significantly faster convergence. Statistically equivalent results marked in bold. Right: Comparison of \ourmethod and MPO for the Cartwheel task both in terms of wall-clock time. A clear advantage for the search based method can be observed.}
    \label{tab:True model results}
\end{figure*}

\paragraph{Results} The performance in terms of wall-clock time after 3h/6h and 48h is presented in Figure \ref{tab:True model results}. We can observe that \ourmethod in combination with the true environment model results in significantly faster convergence than any of the model-free algorithms. 
In addition, as we vary $N$ (the number of rollouts) we observe that for few rollouts \ourmethod behaves similar to a standard RL algorithm. As $N$ increases the tree search starts to find significantly improved policies, resulting in faster convergence. 
Interestingly, the RL methods (which do not make use of an environment model) fail to reach a similar speed-up even when thousands of actors are used.
We speculate that for RL algorithms with an increasing number of actors the policy improvement step (which happens only on the learner) becomes a bottleneck. In contrast, \ourmethod off-loads this step to the actors. With knowledge of the true model actors can locally improve the policy and produce high-quality actions to be consumed by the learner. Note that at the end of learning the high-performing policy will have been distilled into a network and can be executed without reliance on the model.
The training speed-up can thus be interesting in situations where a simulation of the environment is available at training time.

\section{Discussion and Related work}
\label{sec:RelatedWork}
Policy optimization schemes based on the KL regularized objective have a long history in the RL and optimal control literature -- see e.g. \citet{Kappen2005,toussaint2006probabilistic,todorov2007linearly,rawlik12} for different perspectives on this objective. A number of different approaches have been considered for its optimization. These include both policy iteration schemes similar to the ones considered in this paper \citep[e.g.][]{toussaint2006probabilistic,rawlik12,montgomery2016guided,levine2013variational,Theodorou2010,abdolmaleki2018maximum} as well as algorithms that optimize the regularized objective that we consider in the E-step \citep[e.g.][]{ziebart2010modeling,haarnoja2017reinforcement,schulman2017equivalence,hausman2018learning,fox2015taming,haarnoja2018soft}, often via some form of regularized policy gradient. Recently, algorithms in this class have regained considerable attention, including both model-free \citep[e.g.][]{schulman2017equivalence,hausman2018learning,fox2016taming,haarnoja2018soft,maddison2017particle,abdolmaleki2018maximum} and model-based \citep[e.g.][]{levine2013variational,montgomery2016guided,piche2018probabilistic} approaches. In contrast to our work, however, most model-based approaches make use of local dynamics models combined with known reward functions.

Among model-based control for high-dimensional continuous action spaces work by \citet{levine2013variational,montgomery2016guided,chebotar2016path} bears similarity to our work in that they alternate between policy optimization and network fitting, while, for instance, \citet{byravan2019imagined,hafner2019dream} directly use the model to compute model-based policy gradients. Most similar to our work are recent model-based algorithms derived from the perspective of information theoretic optimal control. These optimize for the same objective as our approach but make different assumptions. \citet{piche2018probabilistic} perform planning with a learned model via a form of sequential importance sampling for action selection -- using an separate procedure to optimze a proposal policy. \citet{bhardwaj2019information} use a simulation model to construct K-optimal targets for learning the soft-Q function, as well as during execution for action selection. \citet{lowrey2018plan} rely on the true system dynamics and a learned value function to optimize action sequences during execution. We further expand on the relation to these approaches in the supplementary material.

Monte Carlo tree search \cite{coulom2006efficient} is a well studied family of planning approaches for model-based control. MCTS in combination with learned policies and value functions have recently been successful in challenging problems with discrete action spaces, both with ground-truth 
\citep{silver2017mastering,anthony2017thinking} as well as with learned models \citep{schrittwieser2019mastering}.
There have also been some attempts applying MCTS to problems with continuous action spaces \citep[e.g.][]{yee2016monte,ma2019montecarlo,moerl2018a0c} mostly through an application of the idea of progressive widening \cite{coulom2007computing,chaslot2008progressive} to continuous action spaces \cite{couetouxcontinuous}.
The starting point for our approach, an extension of known regularized policy iteration algorithms to a framework that allows multi-step updates, is quite different; with our tree search effectively being motivated as an approximation to a MC estimate of the soft-Q value.
In light of our positive results it is of course entirely conceivable that other forms of search (perhaps with a clever discretization) will yield similar or even greater benefits.
Exploring such possibilities, and potentially uncovering connections is an exciting direction for future work.

While there has been considerable work on adopting ideas from the probabilistic inference literature to RL, the flow of ideas in the opposite direction has been more limited. One pertinent example is the work by 
\citet{buesing2019approximate} who adapt tree search to perform approximate inference discrete distributions, resulting in a search tree with similar soft-backups to the ones explored in this work.

\section{Conclusion}
We presented a framework for local optimization of the KL-regularized RL objective that allows us to interpolate between model-free and model-based solutions. We explored different algorithm variants and design choices, disentangling benefits of using a model for action selection and policy learning.

Experimentally we show that with a learned model our algorithm achieves a notable improvement in data efficiency compared to state-of-the art model-free approaches. Where a the system model is known (e.g.\ when working with physical simulations) our algorithm allows us to balance computation effectively and can achieve a better computational trade off than conventional high-throughput model-free setups.

Much remains to be done: we have only sampled a small number of design choices within the presented framework, and we expect that the benefits we have observed might transfer to related algorithms. We hope that the perspective of this work will inspire others to investigate other algorithms that flexibly blend the use of model-based and model-free approaches.
\bibliography{references_new}
\bibliographystyle{icml2020}

\appendix

\section{Additional algorithmic details}
We first provide additional details for the derivations of the TreePI procedure from the main paper.

\subsection{K-step optimal policy}
We provide a derivation of the K-step optimal policy from Equation (3) in the main paper. For completeness we re-state the corresponding observation from the main paper:
\begin{observation} 
We can express $\mu^*_{1:K}$ recursively via a K-step soft-optimal Q-function:
\begin{equation}
\begin{aligned}
    \mu^{*}_{K}(a|s) &\propto \pi(a|s) \exp \Big( Q_{\pi}^{*_{K-1}}(s,a) / \alpha \Big) \\
    Q_{\pi}^{*_k}(s,a) &= \Bigg\{\begin{array}{ll}
    r(s) + \gamma \EE_{s' \sim p_{\mu^*_{k}}}[ V^{*_{k-1}}_{\pi}(s') ] &\text{if } k > 1 \\
    r(s) + \gamma \EE_{s' \sim p_{\mu^*_{k}}}[ V^{\pi}_{\pi}(s') ]  &\text{else}
    \end{array} \\
    V^{*_k}_{\pi}(s) &= 
    \alpha \log \int \pi(a|s_t) \exp \Big( Q_{\pi}^{*_k}(s,a) / \alpha \Big) d a. 
\end{aligned}
\end{equation}
\end{observation}
The proof is via induction. 

\textbf{Base case K = 1.} For the base case we have to solve the following maximization problem:
\begin{equation}
\begin{aligned}
   \mu^*_1 &= \arg \max_{\mu_{1}} \objKL^1(\mu_{1}, \pi)  \\ 
   &= \arg \max_{\mu_{1}} \mathbb{E}_{\mu_{1}} \Big [ r_t + \gamma \mathbb{E}_{s' \sim p(s' | s, a)}[V^\pi_\pi(s')] - \alpha KL_t \Big], \\
   &= \arg \max_{\mu_{1}} \mathbb{E}_{\mu_{1}} \Big [ Q^\pi_\pi(s, a) - \alpha KL_t] \\
   &=  \arg \max_{\mu_{1}} \mathbb{E}_{a \sim \mu_{1}} \Big [ Q^\pi_\pi(s, a) - \alpha \log \frac{\mu_1(a | s)}{\pi(a | s)} \Big] \\
   &= \arg \max_{\mu_{1}} \int Q^\pi_{\pi}(s, a) \mu_1(a | s) da \\
   & \ \ \ \ \ \ \ \ \ \ \ \ - \alpha \int \log \frac{\mu_1(a | s)}{\pi(a | s)} \mu_1(a | s) da \\
   &\text{s.t. } \int_a \mu_1(a | s) da = 1
\end{aligned}
\label{eq:max_mu1}
\end{equation}
Forming a Lagrangian for the constraint that $\mu_1$ integrates to one yields
\begin{equation}
\begin{aligned}
\mathcal{L}^1(\mu_1, \lambda)& = \int Q^\pi_{\pi}(s, a) \mu_1(a | s) da  \\
&- \alpha \int \log \frac{\mu_1(a | s)}{\pi(a | s)} \mu_1(a | s) da \\ &+ \lambda (1 - \int_a \mu_1(a | s) da).
\end{aligned}
\end{equation}

Taking the derivative of $\mathcal{L}$ the right hand side wrt. $\mu_1$ ($\nabla_{\mu_1} \mathcal{L}^1(\mu_{1}, \lambda)$) and setting to zero we obtain
\begin{equation}
    0 = Q^\pi_{\pi}(s, a) - \alpha \log \mu_1(a | s) + \alpha \log \pi(a | s) - \lambda.
\label{eq:derivative_mu1}
\end{equation}
Rearranging terms and exponentiating we arrive at 
\begin{equation}
    \mu_1(a | s) = \pi(a | s) \exp\Big(Q^{*_1}_{\pi}(s, a) / \alpha \Big) / \exp(\lambda),
\label{eq:solution_mu1}
\end{equation}
solving for $\lambda$ we obtain the normalizing constant $\lambda = \log Z = \log \int_a \pi(a | s) \exp\Big(Q^{\pi}_{\pi}(s, a) / \alpha \Big)$, with which we finally obtain
\begin{equation}
\mu^*_1(a | s) =  \pi(a | s) \exp\Big(Q^{\pi}_{\pi}(s, a) / \alpha \Big) / Z
\end{equation}
with $Z = \int_a \pi(a | s) \exp\Big(Q^{\pi}_{\pi}(s, a) / \alpha \Big)$ which gives the desired result for $K = 1$. 

Further, expanding the definition of the value function  and inserting $\mu^*_1(a | s)$ we obtain
\begin{equation}
\begin{aligned}
    &V^{*_1}_\pi(s_t) = \mathbb{E}_{\mu_1^*} \Big[  r_t - \alpha KL_t + \gamma \mathbb{E}_{s' \sim p(s'| s, a)}[V^\pi_\pi(s')] \Big] \\
    &= \mathbb{E}_{a \sim \mu_1^*} \Big[  r_t - \alpha \log \frac{\mu^*_1(a | s)}{\pi(a | s)} + \gamma \mathbb{E}_{s' \sim p(s'| s, a)}[V(s')] \Big] \\
    &= \mathbb{E}_{a \sim \mu_1^*} \Big[  Q^\pi_\pi(s, a) - \alpha \log \frac{\mu^*_1(a | s)}{\pi(a | s)} \Big] \\
    &= \mathbb{E}_{a \sim \mu_1^*} \Big[  Q^{\pi}_\pi(s, a) - \alpha \log \frac{\pi(a | s) \exp(Q^{\pi}_{\pi}(s, a)/\alpha) / Z}{\pi(a | s)} \Big] \\
    &= \mathbb{E}_{a \sim \mu_1^*} \Big[  Q^{\pi}_\pi(s, a) - \alpha \log \exp(Q^{\pi}_{\pi}(s, a)/\alpha) + \alpha \log Z \Big] \\
    &= \mathbb{E}_{a \sim \mu_1^*} \Big[ \alpha \log Z \Big] \\
    &= \alpha \log \int \pi(a | s) \exp \Big( Q^\pi_\pi(s, a) / \alpha \Big) da,
\end{aligned}
\label{eq:v_mu1}
\end{equation}
where we have used $\EE_{\mu^*_1} [c] = c$ in the last step. This proves the expected result.

\textbf{Induction from K - 1  to K.} For any $K > 1$ assuming that $\mu^*_{K-1}$ it is easy to see that we can find $\mu^*_K$ via the maximization 
\begin{equation}
\begin{aligned}
    \mu^*_K &= \arg \max_{\mu_K} \objKL^K(\mu_K, \pi) \\
    &= \arg \max_{\mu_{K}} \EE_{\mu_{K}} \Big [ r_t + \gamma \EE_{s' \sim p_{\mu^*_{K}}}[V^{*_{K-1}}_\pi(s')] - \alpha KL_t \Big],
\end{aligned}
\end{equation}
as per Observation \eqref{eq:QMuPiKstep}. We obtain $\mu^*_K(a | s) = \pi(a | s) \exp(Q^{*K-1}_\pi(s, a) \alpha) / Z$ via steps analogous to Equations \eqref{eq:max_mu1}-\eqref{eq:solution_mu1}. From this the definition of $V^{*_K}_\pi(s)$ then immediately follows  analogous to Equation \eqref{eq:v_mu1}. 

\subsection{Derivation of tree estimator}
We give a more detailed explanation of the TreePI estimator (without apprpximations) from Section 4 in the main paper.

Recall that we are interested in a procedure for approximately sampling from $\mu_K^*$. We assume deterministic system dynamics and the availability of a model $s_{t+1} = f(s_{t}, a_{t})$. 

To sample from $\mu_K^*(a | s) \propto \pi(a | s) \exp(Q^{*_K}_\pi(s, a) / \alpha)$ we need an estimate of $\exp(Q^{*_K}_\pi(s, a) / \alpha)$. 
We here show how to obtain such an estimate by building a fully populated tree, of depth $K$. 

\begin{definition}
Let $\hat{Q}^K(s, a)$ be the tree estimate build as follows -- assuming $\alpha = 1$ for brevity of notation. Starting from $s$ we recursively sample $M$ actions per state according to $\pi$, i.e.\ $a_j \sim \pi(\cdot | s)$, up to a depth of $K$ steps. Arrows are labeled with actions. Leaves are set to $Q^\pi_\pi(s,a)$. We can then compute an approximation to $\exp Q^{*_{K}}_\pi(s, a)$ recursively using the approximate value function $\hat{V}^{d+1}(s) = \log \frac{1}{M} \sum_{j=1}^M \exp( r(f(s,a_j)) + \gamma V(f(s,a_j))$ starting from the leaves $\hat{V}^1(s) = \frac{1}{M} \sum_{j=1}^M \exp( Q^\pi_\pi(s,a_j) )$. Then $\hat{Q}^K(s,a) = r(s) + \gamma \hat{V}^{K-1}(f(s,a))$. 
\end{definition}

To prove that this results in the required estimate let us first simplify the derivation: we consider the case of trajectories with finite horizon $|\tau| = T$ and set $\gamma = 1$.  We then seek to prove the following proposition
\begin{observation}
For $\gamma = 1$ and $\alpha = 1$ and deterministic system dynamics the M-sample estimate $\exp \hat{Q}^K(s, a)$ is unbiased. That is $\exp (r(s) + \EE_\pi V^{*_{K-1}}_\pi(s_{t+1}))$= $\EE_\pi[ \exp \hat{Q}^K(s,a)] = \exp Q^{*_{K}}_\pi(s, a)$.
\end{observation}
We start by expanding the definition of $\exp \hat{Q}_\pi^{*_K}(s,a)$, making use of the fact that we consider purely deterministic dynamics of the form $s_{t+1} = f(s_t, a_t)$ and assuming a trajectory of length $T$:
\begin{equation}
\begin{aligned}
    &\exp Q^{*_K}_\pi(s_t, a_t) \\
    = &\exp \Big( r(s_t) + \log \int \pi(a | s_{t+1}) \exp(Q^{*_{K-1}}_\pi(s_{t+1}, a)) da \Big) \\
    = &\exp \Big( r(s_t) \Big)  \int \pi(a | s_{t+1}) \exp(Q^{*_{K-1}}_\pi(s_{t+1}, a)) da \\
    = &\EE_{\pi(a | s_{t+1})} \Big[\exp(r(s_t)) \exp(Q^{*_{K-1}}_\pi(s_{t+1}, a))\Big] \\
    = &\EE_{\pi(a | s_{t+1})} \Big[\exp(r(s_t)) \exp \Big(r(s_{t+1}) \\ 
    & \ \ \ \ \ \ \ \ \ \ \ + \log \int \pi(a' | s_{t+2}) \exp(Q^{*_{K-2}}_\pi(s_{t+2}, a')) da' \Big)\Big] \\
    = &\EE_{\pi(a | s_{t+1})} \Big[\exp(r(s_t)) \exp(r(s_{t+1})) \\ 
    & \ \ \ \ \ \ \ \ \ \ \ \int \pi(a' | s_{t+2}) \exp(Q^{*_{K-2}}_\pi(s_{t+2}, a')) da' \Big] \\
    = &\EE_{a, a' \sim \pi} \Big[\exp(r(s_t)) \exp(r(s_{t+1})) \exp(Q^{*_{K-2}}_\pi(s_{t+2}, a')) \Big] \\
    &\ \ \ \ \ \ \ \ \ \ \ \cdots \\
    = &\EE_{a^{2:K} \sim \pi} \Big[\exp(r(s_t)) \cdots \exp(r(s_{t+K-1})) \\ 
    & \ \ \ \ \ \ \ \ \ \ \ \ \ \  \exp(Q^{\pi}_\pi(s_{t+K}, a^K)) \Big] %\\
    %= &\EE_{a^{2:K} \sim \pi} [\exp Q^{*_K}_\pi(s_t, a_t)],
\end{aligned}
\label{eq:mc_estimate}
\end{equation}
where we recursively expanded the expectation, moving terms outside where possible -- which is possible as the exponential cancels with the logarithm due to the determinism of the environment. Any Monte Carlo approximation to the last line here is easily verified to be an unbiased estimator. This includes our tree based estimate $\exp \hat{Q}^K(s,a)$ in which recursively samples the inner integrals for each state. We note that a corresponding estimate of $V^{*_K}_\pi$ would be biased due to the log in front of the expectation.

It should be noted that this analysis hinges on the fact that $f$ is deterministic and that  $\gamma = 1$ (as is the case for e.g. the humanoid motion capture tracking environments we consider) for infinite horizons and thus $\gamma < 1$ we would not obtain an unbiased estimate. In that case we would instead obtain a pessimistic estimate due to the relation $\exp(\gamma \log \EE_{a \sim \pi}[\exp(f(a))]) \geq \EE_{a \sim \pi}[[\exp(f(a))]^\gamma]$ for $0 < \gamma \leq 1$.

\subsection{Policy iteration procedure}

A full listing of the policy iteration procedure is given in Algorithm \ref{algo:policy_iteration}. 

\begin{algorithm}[tb]
   \caption{K-Step Regularized Policy Iteration}
   \label{algo:policy_iteration}
\begin{algorithmic}
\STATE {\bfseries Input:} number of updates per step $updates\_per\_step$, updates per iteration $target\_rate$  
\STATE {\bfseries Initialize:} $\pi_\theta$, $\hat{Q}^{\pi_\theta}_{\pi_\theta}$, ($r_{\phi_r}$, $f \approx f_\phi$ if needed), $updates = 0$, $i=0$
\REPEAT
\STATE reset $s_t$ to start episode 
\WHILE{not episode ended}
\STATE // sample action with search or $\pi$
\STATE $a \sim \pi^{(i)}(\cdot | s_t)$ or $a \sim$ TreePI (Algorithm 1 in paper)
\STATE // collect next transition
\STATE $s_{t+1} = f(s, a)$, add $(s, a, s_{t+1})$ to buffer $\mathcal{B}$
\FOR{$updates\_per\_step$}
\STATE $s_t = s_{t+1}$
\STATE sample $\tau \in \mathcal{B}$
\STATE update $\hat{Q}^{\pi_\theta}_{\pi_\theta}$, $r_{\phi_r}$, $f_\phi$, $\pi_\theta$
\IF{$updates \mod target\_rate$ {\bfseries equal} 0}
\STATE $\pi^{(i+1)} \leftarrow \pi_\theta$, $\hat{Q}^{\pi^{(i+1)}}_{\pi^{(i+1)}} \leftarrow \hat{Q}^{\pi_\theta}_{\pi_\theta}$
\STATE $i = i + 1$
\ENDIF
\STATE $updates \mathrel{+}=  num\_update\_steps$
\ENDFOR
\ENDWHILE
\UNTIL{convergence}
\end{algorithmic}
\end{algorithm}

\section{Additional details on the model setup}

\subsection{Hyperparameters for \ourmethod}
We use fully connected neural networks with
the following structure: 
If the model is learned then the state-encoder $f_{\phi_\text{enc}}$ consist of two fully-connected layers with 256 units each with exponentiated linear (elu) activation functions \citep{eluClevert15}, layer normalization is used in the first layer of the network (which we find useful for dealing with different observation scales accross domains). The transition model similarly contains two layers of 256 units and predicts a delta that is added to the previous state (a standard choice that simplifies the learning problem). The latent state size is assumed to be 256. To perform forward prediction we feed the transition model with both the encoder state (or previous predicted state) as well as the action -- which simply concatenate to the latent features.

The reward, Q and policy networks operate either on the ground truth states (when a model of the system is available, i.e. in the humanoid experiments from Section 6.2) or on the shared representation obtained from $f_{\phi_\text{enc}}$ and  $f_{\phi_\text{trans}}$.
The networks each contain three layers of 256 units each (with layer normalization after the first layer); followed by a linear layer outputting scalar values for $r_{\phi_r}, \hat{Q}_{\phi_Q}$ and two layers for prediction policy mean $\mu_{\theta}(x)$ and log variance $\varsigma_\theta(x)$ of the policy for $\pi_\theta(a | s) = \mathcal{N}(\mu_{\theta}(x), I \text{softplus}(\varsigma_\theta(x)))$ respectively. For the Q-function the action is concatenated to the other inputs.

We tuned other hyperparameters such as $M$ and $\alpha$ on the walker domain and then fixed them for the other experiments.
All parameters are listed in Table \ref{t:hypers_treepi}. The algorithm was not overly sensitive to the exact settings of the parameters as long as they were in the right order of magnitude.

\begin{table}[t]
\begin{center}
 \begin{tabular}{c||c} 
 Hyperparameters & TreePI \\
 \hline
 Encoder $f_\text{enc}$ & 256-LN-256\\ 
 Transition $f_\text{trans}$ & 256-LN-256\\
 Policy net & 256-LN-256-256\\ 
 Q function network & 256-LN-256-256\\
 Branching factor M & 20\\
 Maximum Depth K & 10\\
 Number of rollouts N & 100\\
 $\alpha$ & 0.1 \\
 $\epsilon_\text{KL}$ & 0.005 \\
 Discount factor ($\gamma$) & 0.99 \\
 Adam learning rate & 0.0003 \\
 Replay buffer size & 2000000 \\
 Target network update period & 200\\
 Batch size & 256\\
 Activation function & elu\\
\end{tabular}
\end{center}
\caption{Hyperparameters for TreePI. LN corresponds to layer-normalization. }
\label{t:hypers_treepi}
\end{table}

\subsection{Hyperparameters for the baselines}

For the baseline experiments we constructed policy and Q-function networks that are aligned with the above described network architecture. For \textbf{MPO+Predictive} we use the exact same network architecture as outlined in Table \ref{t:hypers_treepi}. For the other baselines (\textbf{MPO}, \textbf{MPO}, \textbf{PPO}) we construct a network without th predictive parts -- that is otherwise similar to the TreePI setup. Concretely, we apply an encoder $f_\text{enc}$ to the provided obervations, followed by networks predicting policy parameters and Q-function (or Value function in the case of PPO).
The hyperparameters of individual algorithms are listed in Table \ref{t:hypers_other}

\begin{table}[t]
\begin{center}
 \begin{tabular}{c||c} 
 Hyperparameters & All methods \\
 \hline
 Encoder $f_\text{enc}$ & 256-LN-256\\ 
 Policy net & 256-LN-256-256\\
 Activation function & elu\\
 Discount factor ($\gamma$) & 0.99 \\
 \hline
 MPO & \\
 \hline
 Q function network & 256-LN-256-256\\
 $\epsilon_\text{E-Step}$ & 0.1 \\
 $\epsilon_\text{mean}$ & 0.005 \\
 $\epsilon_\text{cov}$ & 0.0001 \\
 Adam learning rate & 0.0003 \\
 Replay buffer size & 2000000 \\
 Target network update period & 200\\
 Batch size & 256\\
 \hline
 SAC & \\
 \hline
 Q function network & 256-LN-256-256\\
 Target entropy (per action dimension) & -0.5 \\
 Adam learning rate & 0.0003 \\
 Replay buffer size & 2000000 \\
 Target network update period & 200\\
 Batch size & 512\\
 \hline
 PPO & \\
 \hline
 Value function network & 256-LN-256-256\\
 Clipping $\epsilon$ & 0.2 \\
 Adam learning rate & 0.0002 \\
 Batch size & 512 \\
 Sequence length & 10 \\
\end{tabular}
\end{center}
\caption{Hyperparameters for TreePI. LN corresponds to layer-normalization. }
\label{t:hypers_other}
\end{table}

In the distributed setting multiple actors were used to generate experience for the learner. Each actor has generates experience by performing an episode in the environment, data is sent to the learner after each executed transition (either filling a replay buffer or a queue (for PPO)). Parameters are then fetched from the learner every after 100 interactions with the environment.

\section{Additional details on the Policy Gradient Variant}
As described in the main text, we performed an additional ablation where we replaced the search at action selection time with a policy gradient that temporarily changes the network parameters.
We now derive the policy gradient used for this experiment. We start by recalling the K-step objective
\begin{equation}
    \objKL^K(\mu, \pi) = \EE_{\mu_{1:K}} \Big [ \gamma^{K} V^\pi_\pi(s_{K+1}) + \sum_{t=1}^{K} \gamma^{t - 1} (r_{t} - \alpha KL_{t}) \Big], \label{eq:objective_k_step}
\end{equation}
where, as before, $KL_{t} = \KL[\qp(\cdot | s_t) \| \pp(\cdot | s_t)]$.
We now consider parametric policy $\mu_{\theta_\mu}$ initialized with the parameters of $\pi^{(i)}_\theta$. 
Defining the KL regularized K-step return as
$$
R = \gamma^{K} V^\pi_\pi(s_{K+1}) + \sum_{t'=t}^{K} \gamma^{t' - 1} (r_{t'} - \alpha KL_{t'})
$$ 
and the corresponding return from time $t$ onwards as $$
R_{t:K} = \gamma^{K} V^\pi_\pi(s_{K+1}) + \sum_{t'=t}^{K} \gamma^{t' - 1} (r_{t'} - \alpha KL_{t'}),
$$
we can then write the required derivative as
\begin{equation}
\begin{aligned}
    &\nabla_{\theta_\mu} \objKL(\mu_{\theta_\mu}, \pi^{(i)}_\theta) \\ 
    &= \nabla_{\theta_\mu} \EE_{\mu_{\theta_\mu}} \Big [ \gamma^{K} V^\pi_\pi(s_{K+1}) + \sum_{t=1}^{K} \gamma^{t - 1} (r_{t} - \alpha KL_{t}) \Big] \\
    &= \EE_{\mu_{\theta_\mu}} \Big [ \sum_{t=1}^K R \nabla_{\theta_\mu} \log \mu_{\theta_\mu}(a_t | s_t) \Big] \\
    &= \EE_{\mu_{\theta_\mu}} \Big [ \sum_{t=1}^K R_{t:K} \nabla_{\theta_\mu} \log \mu_{\theta_\mu}(a_t | s_t) \Big],
\end{aligned}
\label{eq:pg}
\end{equation}
where from the second to third line we have made use of the standard likelihood ratio policy gradient \citep{williams1992simple} and from the third to fourth line dropped terms not influenced by the decision at decision point $s_t$ (i.e. the action choice at t does not influence the past rewards). 

Equation \ref{eq:pg} can be estimated by performing rollouts (against the learned model $f$). To change the policy parameters at state $s_t$ during policy execution we then perform 10 gradient steps (using Adam as the optimizer with learning rate $0.0005$) estimating the gradient in each step based on 100 rollouts (of depth 10). After these 10 updates we sample an action from $\mu_{\theta_\mu}$ and reset its parameters to $\theta$.  This procedure effectively replaces the action sampling step in Algorithm 1. 

\section{Additional details on relations to existing algorithms}
As mentioned in the main paper, in the context of EM policy iteration schemes \citet{abdolmaleki2018maximum} implement the same estimate for $\mu$ as our algorithm if we set $K=1$. In this setting the main difference is how $\alpha$ is set. While we treat $\alpha$ as a hyper-parameter, MPO starts from a hard-constraint (instead of the KL regularization we employ), performs Lagrangian relaxation to obtain a dual-function and then optimizes this dual for a target maximum KL ($\epsilon_{\text{E-step}} = 0.1$). 

Additionally, if we keep $K > 1$ but drop the nested optimization wrt. $\mu_1, \mu_2, \dots, \mu_{K-1}$ we obtain a different approximate solution to the optimal policy $\mu^*$ given as 
\begin{equation}
\mu^\text{PI}_K(a | s_t) \propto \pi(a | s_t) \exp\big(\gamma^K V^\pi_\pi(s_{t+K}) + \nicefrac{\sum_{t'} \gamma^{t'-1} r_t}{\alpha}\big),
\end{equation} which corresponds to estimating the K-step optimal path  at state $s_t$ (as estimated via a K-step rollout following $\pi$) assuming that $\pi$ is followed thereafter. This solution can be realized via a simple Monte Carlo estimator (to estimate the path costs in the exponential, for different actions at $s_t$) and can be related to the path-integral formulation for policy improvement \citep{Kappen2005,Theodorou2010}. Concretely, considering $\mu^\text{PI}_K$ corresponds to a path integral with $\pi$ acting as the prior and (approximate) boot-strapping according to the value function $V^\mu_\pi$. As mentioned, we can realize this improved policy by sampling paths simulated by following $\pi$ and then re-weighting with the exponentiated return. An interesting variant would be to target the same density $\mu^\text{PI}_K$ with a more advanced sampler such as a Sequential Monte Carlo (SMC) method -- which starts simulating M paths from $s_t$ but re-samples after each step according to the exponentiated accumulated reward. Following such a procedure recovers the sampling employed by \citet{piche2018probabilistic}. 

Overall, Our policy $\pi_{1:K}(a | s)$ from Eq. \eqref{eq:QMuPiKstep} can thus be seen as an extension of path-integral control where we consider tree-like branching in each state -- up to depth K -- rather than linear paths to estimate an improved policy. 

\section{Additional details on the humanoid Motion Capture domain}
We give additional details on the experiments for reproducing motion capture data. In these experiments the observations consist of the humanoid's proprioceptive state (given by joint positions and velocities) as well as the head position in Euclidean space and time (encoded via a phase variable that starts at 0 at the beginning of an episode and is 1 at the end of the episode).

The setup is thus similar to \citep{merel2018hierarchical} and we use the same complex humanoid body.
We note that, Deep RL approaches are currently popular for producing controllers that track motion-capture data but the training process can be quite slow \citep[e.g.][]{peng2018deepmimic}. Earlier
works have used model-based search to control humanoids \citep{hamalainen2014online} and match reference movements \citep{liu2010sampling}.

We use a time-varying reward that measures the discrepancy between the body-pose of the humanoid and the a reference pose from a motion capture clip. We 
and train a policy to reproduce a cartwheel and backflip motion extracted from the CMU Motion Capture Database\footnote{http://mocap.cs.cmu.edu/} (see the supplementary for a full description of the task). 
The reward function we use is given as
\begin{equation}
    r_t = \exp(-\beta \text{E}_\text{total} / w_\text{total}),
\end{equation}
where $\text{E}_\text{total}$ is a sum of deviations from the reference trajectory and $w_\text{total}$ a sum of corresponding weights the total sum of deviations is given as 
\begin{equation}
\begin{aligned}
\text{E}_\text{total} = &w_\text{vel}  D(q_\text{vel}, q^*_\text{vel}, N_\text{vel}) + w_\text{pos} D(q_\text{pos}, q^*_\text{pos}, N_\text{pos}) \\
 +&w_\text{ori} \| \log(q_\text{ori} * (q^*_\text{ori})^{-1}) \| + w_\text{ee} D(q_\text{ee}, q^*_\text{ee}, N_\text{ee}) \\
 & + w_\text{xvel} D(q_\text{xvel}, q^*_\text{xvel}, N_\text{xvel}),
\end{aligned}
\end{equation}
where $D(x, y, N) = \nicefrac{1}{N} \sum | x - y |$, and where vel and pos refer to velocity and position componjents of the joint positions respectively, xvel and xpos are the velocity and position in cartesian space and ee denotes the end-effector positions of the humanoids body. 
We use $w_\text{vel} = 1$, $w_\text{pos} = 5$, $w_\text{ori} = 20$, $w_\text{ee} = 1$, $w_\text{xvel} = 1$.

\section{Additional comparison to SMC at action selection time}
\begin{figure}
    \centering
    \includegraphics[width=\columnwidth]{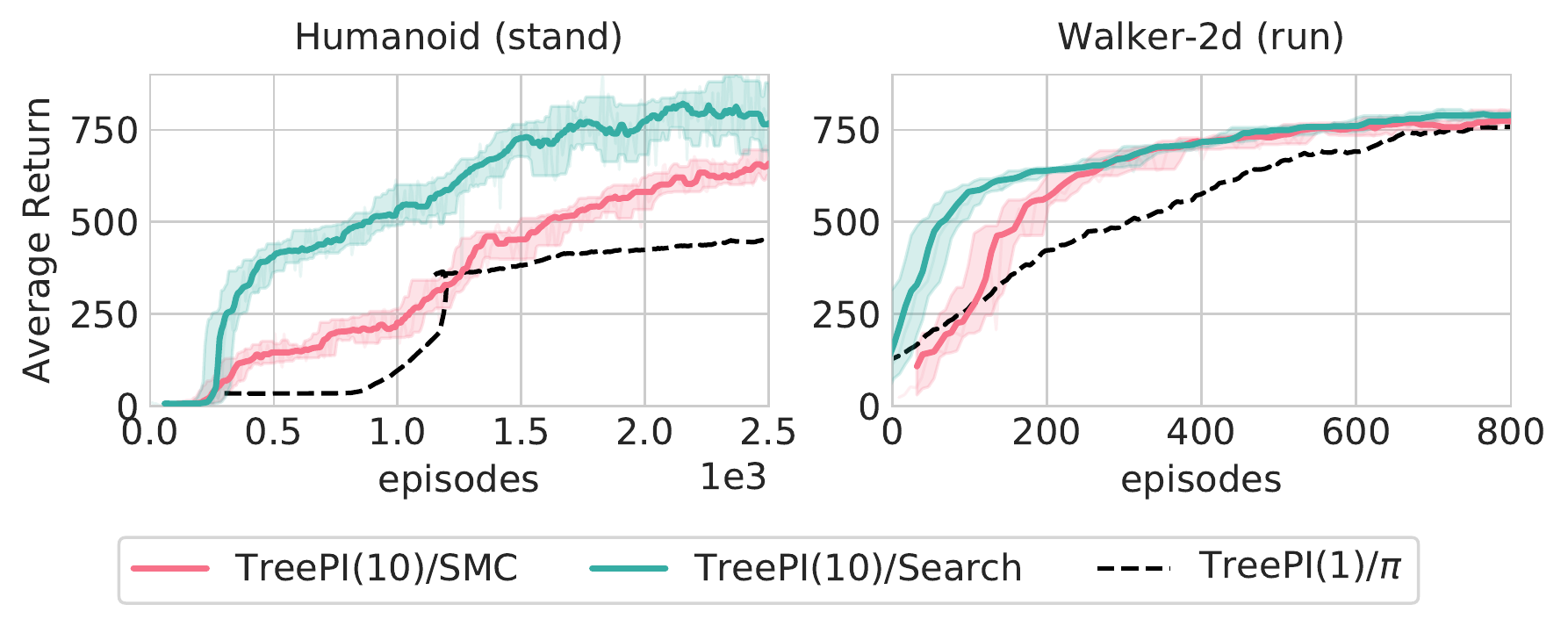}
    \caption{Comparison to sampling via SMC instead of using search via TreePI at action selection time.}
    \label{fig:compare_smc}
\end{figure}

As an additional ablation that was not contained in the main paper -- and to support the hypothesis that search at action selection time can generally be useful -- we pair TreePI with another approach for sampling approximately from $\mu^*_K$. Specifically we make use of a Sequential Monte Carlo sampler (as also discussed in Section 4 of the supplementary), akin to \citet{piche2018probabilistic}, but here paired with our TreePI method for optimizing the policy during the optimization.
The results are shown in Figure \ref{fig:compare_smc}. We observe that, similar to the policy gradient ablation in the main paper, a significant benefit can be obtained over Treepi/$\pi$.

\end{document}